\documentclass{article}

\usepackage{amsmath}
\usepackage{amssymb}
\usepackage{mathtools}
\usepackage{amsthm}
\usepackage{float}
\usepackage{hyperref}
\usepackage{url}
\usepackage{amsfonts}
\usepackage{array}
\usepackage[caption=false,font=normalsize,labelfont=sf,textfont=sf]{subfig}
\usepackage{textcomp}
\usepackage{stfloats}
\usepackage{verbatim}
\usepackage{graphicx}
\usepackage{booktabs}
\usepackage{color}
\usepackage{multirow}
\usepackage{enumerate}
\usepackage{bbding}
\usepackage{wrapfig}

\usepackage[accepted]{draft}

\usepackage{amsmath}
\usepackage{amssymb}
\usepackage{mathtools}
\usepackage{amsthm}
\usepackage{caption}
\usepackage{bm}

\usepackage[capitalize,noabbrev]{cleveref}

\theoremstyle{plain}

\theoremstyle{definition}

\theoremstyle{remark}

\usepackage[textsize=tiny]{todonotes}

\icmltitlerunning{}

\begin{document}

\twocolumn[
\icmltitle{GaussianPro: 3D Gaussian Splatting with Progressive Propagation}




\icmlsetsymbol{equal}{*}

\begin{icmlauthorlist}
\icmlauthor{Kai Cheng}{equal,sch1}
\icmlauthor{Xiaoxiao Long}{equal,sch2}
\icmlauthor{Kaizhi Yang}{sch1}
\icmlauthor{Yao Yao}{sch3}
\icmlauthor{Wei Yin}{comp}
\icmlauthor{Yuexin Ma}{sch4}
\icmlauthor{Wenping Wang}{sch5}
\icmlauthor{Xuejin Chen}{sch1}
\end{icmlauthorlist}

\icmlaffiliation{sch1}{University of Science and Technology of China}
\icmlaffiliation{sch2}{The University of Hong Kong}
\icmlaffiliation{sch3}{Nanjing University}
\icmlaffiliation{comp}{The University of Adelaide}
\icmlaffiliation{sch4}{ShanghaiTech University}
\icmlaffiliation{sch5}{Texas A\&M University}

\icmlcorrespondingauthor{Xuejin Chen}{xjchen99@ustc.edu.cn}

\icmlkeywords{Machine Learning, ICML}

\vskip 0.3in

\renewcommand\twocolumn[1][]{#1}%
\vspace{-6mm}%
\begin{center}
    \centering
	\includegraphics[width=\textwidth]{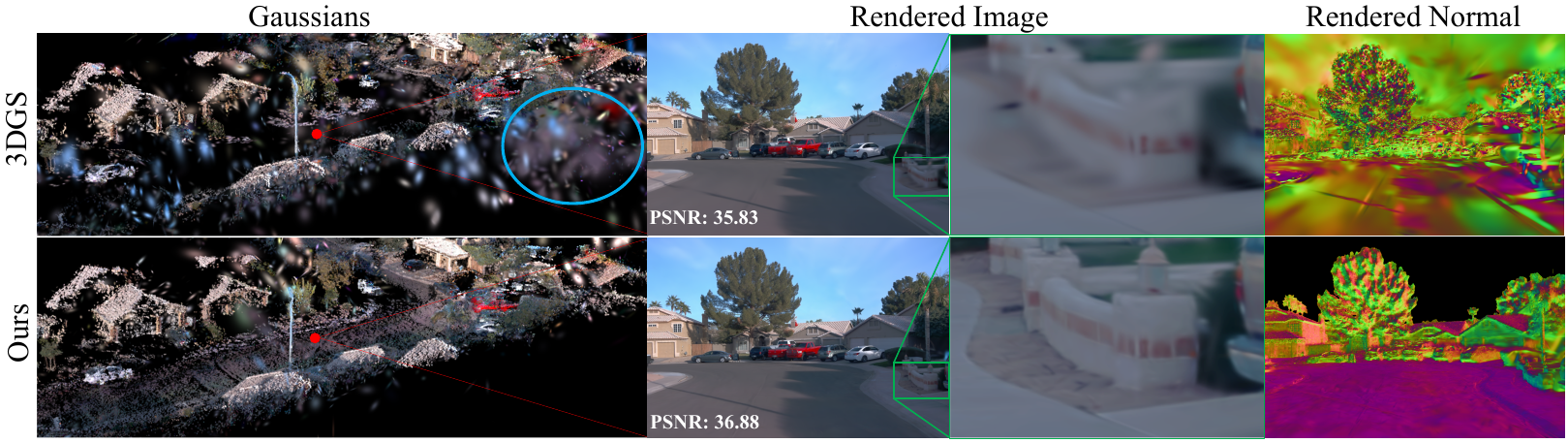}
	\vspace{-6mm}
	\captionof{figure}{The sparse SfM points and less-constrained densification strategies of 3DGS pose challenges in optimizing 3D Gaussians, particularly for textureless areas. 3DGS generates incorrect Gaussians (blue circle) to be over-fitted on the training images, leading to a noticeable performance drop in novel view rendering with erroneous geometries.}
    \label{fig:teaser}
    \vspace{2mm}%
\end{center}%

]



\printAffiliationsAndNotice{\icmlEqualContribution} 


\begin{abstract}

The advent of 3D Gaussian Splatting (3DGS) has recently brought about a revolution in the field of neural rendering, facilitating high-quality renderings at real-time speed. However, 3DGS heavily depends on the initialized point cloud produced by Structure-from-Motion (SfM) techniques.
When tackling with large-scale scenes that unavoidably contain texture-less surfaces, the SfM techniques always fail to produce enough points in these surfaces and cannot provide good initialization for 3DGS. As a result, 3DGS suffers from difficult optimization and low-quality renderings.
In this paper, inspired by classical multi-view stereo (MVS) techniques, we propose GaussianPro, a novel method that applies a progressive propagation strategy to guide the densification of the 3D Gaussians. 
Compared to the simple split and clone strategies used in 3DGS, our method leverages the priors of the existing reconstructed geometries of the scene and patch matching techniques to produce new Gaussians with accurate positions and orientations.
Experiments on both large-scale and small-scale scenes validate the effectiveness of our method, where our method significantly surpasses 3DGS on the Waymo dataset, exhibiting an improvement of 1.15dB in terms of PSNR.

\end{abstract}

\section{Introduction}

Novel view synthesis is an important but challenging task in computer vision and computer graphics that aims to generate images of novel viewpoints in the captured scene. 
It has extensive applications in various domains, including virtual reality~\cite{deng2022fov}, autonomous driving~\cite{yang2023unisim,cheng2023uc}, and 3D content generation~\cite{poole2022dreamfusion,tang2023dreamgaussian}.
Recently, Neural Radiance Fields (NeRF)~\cite{mildenhall2020nerf} has significantly boosted this task, achieving high-fidelity renderings without explicitly modeling 3D scenes, texture and illumination. 
However, due to the heavy manner of volume rendering, NeRFs still suffer from slow rendering speed, although various efforts~\cite{muller2022instant, barron2022mip, barron2023zip, chen2022tensorf, xu2022point} have been made. 


To achieve real-time neural rendering, 3D Gaussian Splatting (3DGS)~\cite{kerbl20233d} has been developed. It models the scenes explicitly as 3D Gaussians with learnable attributes and performs rasterization of the Gaussians to produce renderings.
The splatting strategy avoids time-consuming ray sampling and allows parallel computations, thus yielding high efficiency and fast rendering. 
3DGS heavily relies on the sparse point clouds produced by Structure-from-Motion (SfM) techniques to initialize the Gaussians, e.g., their positions, colors, and shapes. Moreover, clone and split strategies to create more new Gaussians to achieve a complete coverage of the scene.
However, the densification strategies to create 3D Gaussians lead to two main limitations.
1) \textbf{Sensitive to Gaussian Initialization}. The SfM techniques always fail to produce 3D points and leave empty in textureless regions, and therefore the densification strategy struggles to generate reliable Gaussians to cover the scene with a poor initialization.
2) \textbf{Ignore the priors of the existing reconstructed geometry.} The new Gaussians are either cloned to be the same as the old Gaussians or initialized with random positions and orientations. 
The less-constrained densification leads to difficulties in the optimization of 3D Gaussians, e.g., noisy geometries, and few Gaussians in texture-less regions, finally degrading the rendering quality.
As shown in Figure~\ref{fig:teaser}, the results of 3DGS contain many noisy Gaussians and some regions are not covered by enough Gausians.

In this paper, we propose a novel progressive propagation strategy to facilitate 3DGS, which could produce more compact and accurate 3D Gaussians and therefore improve the rendering quality, especially in texture-less surfaces.
The key idea of our method is to fully leverage the reconstructed scene geometries as priors and classical patch matching techniques to progressively produce new Gaussians with accurate positions and orientations.

Specifically, we consider Gaussian densification in both 3D world space and 2D image space. For each input image, we render the depth and normal map by accumulating the positions and orientations of 3D Gaussians via alpha blending. 
Based on the observation that the neighboring pixels are likely to share similar depth and normal values, for a pixel, we iteratively propagate the depth and normal values of its neighboring pixels to this pixel to formulate a set of candidates. 
With the candidates, we leverage classical patch matching techniques to pick up the best candidate that satisfies the multi-view photometric consistency constraint, thus yielding new depth and normal for each pixel (named as propagated depth/normal). 
We select the pixels whose propagated depth is significantly different from the rendered depth since large differences imply that the existing 3D Gaussians may not accurately capture the true geometry. 
As a result, we explicitly back-project the selected pixels using the propagated depths into 3D space and initialize them as new Gaussians.
Additionally, we leverage the propagated normals to regularize the orientations of 3D Gaussians, further improving the reconstructed 3D geometry and rendering quality.


Our proposed progressive propagation strategy could produce more compact and accurate 3D Gaussians by transferring of accurate geometric information from well-modeled regions to under-modeled regions. 
As shown in Figure.~\ref{fig:teaser}, compared to 3DGS, our method produces more accurate and compact Gaussians and therefore achieves a better coverage of the 3D scene.
Experiments on public datasets such as Waymo and MipNeRF360 validate that our proposed strategy significantly boosts the performance of 3DGS.
Overall, the contributions of our method are summarized as:
\begin{itemize}
  \item We propose a novel Gaussian propagation strategy that guides the densification to produce more compact and accurate Gaussians, particularly in low-texture regions.
  \item We additionally leverage a planar loss that provides a further constraint in the optimization of Gaussians.
  \item Our method achieves new state-of-the-art rendering performance on the Waymo and MipNeRF360 datasets. Our method also presents robustness to the varying numbers of input images.
\end{itemize}

\section{Related Work}
\subsection{Multi-view Stereo}
Multi-view stereo (MVS) aims to reconstruct a 3D model from a collection of posed images, which can be further combined with traditional rendering algorithms to generate novel views. Traditional methods~\cite{campbell2008using, furukawa2009accurate, bleyer2011patchmatch, furukawa2015multi, schonberger2016pixelwise, xu2019multi} explicitly establish pixel correspondences between images based on hand-crafted image features and then optimize the 3D structure to achieve the best pixel correspondences among images. Learning-based MVS methods~\cite{yao2018mvsnet, vakalopoulou2018atlasnet, long2020occlusion, chen2019point, long2021multi, ma2022multiview, long2022sparseneus, feng2023cvrecon} implicitly build multi-view correspondences with learnable features and regress depth or 3D volume based on the features in an end-to-end framework. In this paper, we draw inspiration from depth optimization in MVS to improve the geometry of the Gaussians, thereby achieving better rendering results.

\subsection{Neural Radiance Field}
NeRF combines deep learning techniques with the 3D volumetric representation, transforming a 3D scene into a learnable continuous density field. Utilizing ray marching in volume rendering, NeRF is able to achieve high-quality novel view synthesis without explicit modeling of the 3D scene and illumination. 
To further improve the rendering quality, some approaches~\cite{barron2021mip, xu2022point,barron2023zip} directly improve the point sampling strategy in ray marching for more accurate modeling of the volume rendering process. Others~\cite{barron2022mip, wang2023f2} improve rendering by reparameterizing the scene to generate more compact scene representation and easier learning process. Additionally, regularization terms~\cite{deng2022depth, yu2022monosdf} could be introduced to constrain the scene representation towards a closer approximation of real geometry. Despite these advancements, NeRF still incurs high computational costs during rendering. 
Since NeRF employs MLPs to represent the scene, the computation and optimization of any point in the scene are dependent on the entire MLP. Many works propose novel scene representations to accelerate rendering. They replace MLPs with sparse voxels~\cite{liu2020neural, fridovich2022plenoxels}, hash tables~\cite{muller2022instant}, or triplane~\cite{chen2022tensorf}, allowing the computation and optimization of each point to be localized to the corresponding local region of the scene.
Although these methods significantly improve rendering speed, real-time rendering is still challenging due to the inherent ray marching strategy in volume rendering.

\subsection{3D Gaussian Splatting}
3DGS employs a spatting-based rasterization~\cite{zwicker2002ewa} approach to project anisotropic 3D Gaussians onto a 2D screen. It computes the pixel's color by performing depth sorting and $\alpha$-blending on the projected 2D Gaussians, which avoids the sophisticated sampling strategy of ray marching and achieves real-time rendering. Some concurrent works have made improvements to 3DGS. Firstly, 3DGS is sensitive to sampling frequency, i.e., changing the camera's focal length or camera distance could result in rendering artifacts. These artifacts are addressed by introducing low-pass filtering~\cite{yu2023mip} or multi-scale Gaussian representations~\cite{yan2023multi}. Additionally, 3DGS excessively grows Gaussians without explicitly constraining the scene's real geometric structure, resulting in numerous redundant Gaussians and significant memory consumption.
Some methods evaluate the contribution of Gaussians to rendering by their scales~\cite{lee2023compact} or calculating their visibility in views~\cite{fan2023lightgaussian}, forcing the removal of Gaussians with small contributions. Others compress the storage of Gaussian attributes by quantization technique~\cite{navaneet2023compact3d} or interpolating Gaussian attributes from structured grid features~\cite{morgenstern2023compact, lu2023scaffold}.

Although these methods significantly reduce the storage overhead of Gaussians, they do not explicitly constrain the geometry of the Gaussians.
3DGS could grow in locations far from the real surfaces to fit different training views, resulting in redundancy and a decrease in rendering quality for new viewpoints. This paper considers the planar prior in the scene, explicitly constraining the growth of Gaussians close to the real surfaces. This approach enables Gaussians to better fit the real geometry of the scene, achieving improved rendering and more compact representation.

\section{Preliminaries}
3DGS~\cite{kerbl20233d} models the 3D scene as a set of anisotropic 3D Guassians, which are further rendered to images using the splatting-based rasterization technique~\cite{zwicker2002ewa}. 
For each 3D Gaussian $G$, it is defined as:
\begin{equation}
\label{eq:gaussian}
G(\mathbf{x})=e^{-\frac{1}{2}(\mathbf{x}-\bm{\mu})^T \bm{\Sigma}^{-1}(\mathbf{x}-\bm{\mu})},
\end{equation}
where $\bm{\mu} \in \mathbb{R}^{3 \times 1}$ refers to its mean vector, $\bm{\Sigma} \in \mathbb{R}^{3 \times 3}$ refers to its covariance matrix. In order to ensure the positive semi-definite property of the covariance matrix during the optimization, it is further expressed as $\bm{\Sigma}=\mathbf{R} \mathbf{S} \mathbf{S}^T \mathbf{R}^T$, where the rotation matrix $\mathbf{R} \in \mathbb{R}^{3 \times 3}$ is orthogonal, and the scale matrix $\mathbf{S} \in \mathbb{R}^{3 \times 3}$ is diagonal. 

To render an image from a given viewpoint, the color of each pixel $\mathbf{p}$ is calculated by blending $N$ ordered Gaussians $\left\{G_i \mid i=1, \cdots ,N\right\}$ overlapping $\mathbf{p}$ as 
\begin{equation}
\label{eq:colorblending}
\mathbf{c}(\mathbf{p})=\sum_{i=1}^N \mathbf{c}_i \alpha_i \prod_{j=1}^{i-1}\left(1-\alpha_j\right),
\end{equation}
where $\alpha_i$ is obtained by evaluating a projected 2D Gaussian~\cite{zwicker2002ewa} from $G_i$ in $\mathbf{p}$ multiplied with a
learned opacity of $G_i$, and $\mathbf{c}_{i}$ is the learnable color of $G_i$. Gaussians that cover $\mathbf{p}$ are sorted in ascending order of their depths under the current viewpoint. Through differentiable rendering techniques, all attributes of the Gaussians could be optimized end-to-end via training view reconstruction.



In order to accurately represent the scene geometry, 3DGS also employs a densification strategy to generate new Gaussians. For each training iteration, if the gradient backpropagated from the rendering loss to the current Gaussian exceeds a certain threshold, 3DGS considers that it does not sufficiently represent the corresponding 3D region. If the covariance of the Gaussian is large, it is split into two Gaussians. Conversely, if the covariance is small, it is cloned. This strategy encourages 3DGS to increase the number of Gaussians to cover the captured scene.

\section{Method}

\begin{figure*}
\centering    
\includegraphics[width=\linewidth]{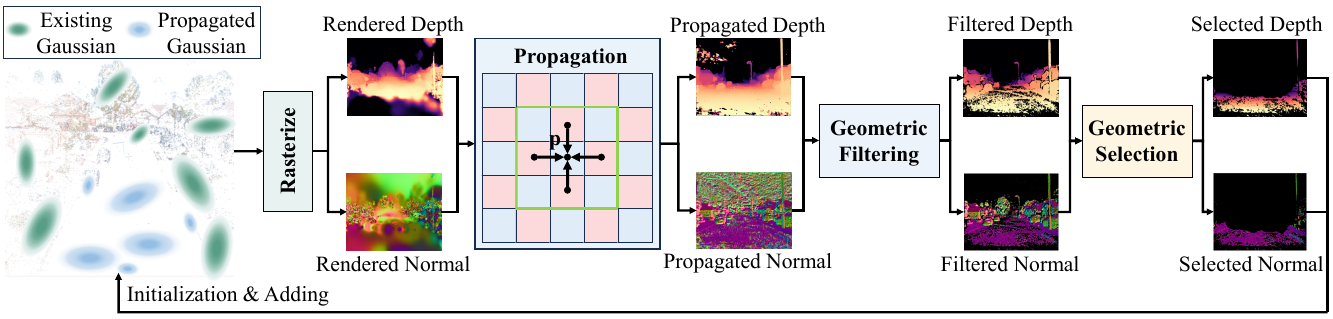}
 \caption{Progressive Propagation of Gaussian. Firstly, we render the depth and normal maps from the 3D Gaussians. Then we iteratively perform propagation operations on the rendered depths and normals to generate new depth and normal values (denoted as propagated depth and propagated normal) via patch matching techniques.
 We filter out the unreliable propagated depths and normals using geometric consistency, yielding filtered depths and filtered normals. Finally, we identify the regions where their rendered depths and normals significantly deviate from the filtered ones, indicating that existing Gaussians may not inaccurately capture the geometry and therefore need more Gaussians. Pixels in these regions are projected into the 3D space to initialize new Gaussians using the filtered depth and normal.} 
\label{fig: propagation}
\end{figure*}

\subsection{Overview}
In this paper, we propose a novel progressive propagation strategy to explicitly generate 3D Gaussians with accurate positions and orientations, thereby improving rendering quality and compactness. 
First, instead of only coupling with 3D space, we propose to tackle this problem in both 3D space and 2D image space.
We project 3D Gaussians onto the 2D space to generate depth and normal maps, which are used to guide the growth of Gaussians (Sec.~\ref{sec:2dmapping}).
Then we iteratively update each pixel's depth and normal based on the propagated ones from its neighboring pixels. Pixels whose new depth is significantly different from the initial depth are projected back to 3D space as 3D points and these points are further initialized as new Gaussians (Sec.\ref{sec:propagation}). Additionally, a planar loss is also incorporated to further regularize the geometry of the Gaussians, yielding more accurate geometries (Sec.~\ref{sec:normal}). The overall training strategy is introduced in Sec.~\ref{sec:training}.

\subsection{Hybrid Geometric Representation}
\label{sec:2dmapping}
In this section, we propose a hybrid Geometric representation that combines 3D Gaussians with 2D view-dependent depth and normal maps, where the 2D representations are utilized to assist the densification of Gaussians. 

Due to the discrete and irregular topology of the 3D Gaussians, 
it is inconvenient to perceive the connectivity of geometries, like searching neighboring Gaussians on a local surface. As a result, it's difficult to perceive the existing geometry to guide the Gaussian densification.
Inspired by the classical MVS methods, we propose to tackle this challenge by mapping the 3D Gaussians into structured 2D image space. This mapping allows us to efficiently determine the neighbors of the Gaussians and propagate geometric information among them.
Specifically, when Gaussians are located on the same local plane in 3D space, their 2D projections should also be in adjacent regions and exhibit similar geometric properties, i.e. depth and normal. 

\textbf{The depth value of Gaussian.}
For each viewpoint with camera extrinsics $[\mathbf{W}, \mathbf{t}] \in \mathbb{R}^{3 \times 4}$, the center $\bm{\mu}_i$ of a Gaussian $G_i$ can be projected into the camera coordinate system as $\bm{\mu}_i^{\prime}$:
\begin{equation}
\bm{\mu}_{i}^{\prime}=\left[\begin{array}{c}
x_{i} \\
y_{i} \\
z_{i}
\end{array}\right]=\mathbf{W} \bm{\mu}_{i}+\mathbf{t},
\end{equation}
where $z_i$ refers to the Gaussian's depth under the current viewpoint.

\textbf{The normal value of Gaussian.}
In a Gaussian $G_i$, its covariance matrix is formulated as $\bm{\Sigma}_i=\mathbf{R}_i \mathbf{S}_i \mathbf{S}_i^T \mathbf{R}_i^T$. The rotation matrix $\mathbf{R}_i$ determines its three orthogonal eigenvectors while scaling matrix $\mathbf{S}_i \in \mathbb{R}^{3 \times 3}$ determines the scale along the eigenvector directions. The covariance matrix $\bm{\Sigma}_{i}$ of a 3D Gaussian could be compared to a representation of the shape of an ellipsoid, where the eigenvectors correspond to ellipsoid's axes and the scales refer to the lengths of the axes.
According to the GaussianShader~\cite{jiang2023gaussianshader}, the Gaussian sphere gradually becomes flattened and approaches a plane during the optimization process. Therefore, the direction of its shortest axis can approximate the normal direction $\mathbf{n}_i$ of the Gaussian, which is induced by
\begin{equation}
\mathbf{n}_i=\mathbf{R}_i[r, :], r=\operatorname{argmin}\left(\left[s_1, s_2, s_3\right]\right),
\end{equation}
where $diag(s_1, s_2, s_3) = \mathbf{S}_i$, $\operatorname{argmin}(\cdot)$ is the operation to find the index of the minimum value.

Finally, the 2D depth and normal map under the current viewpoint are rendered based on $\alpha$-blending defined in Eq.~\ref{eq:colorblending}, where the attribute color $\mathbf{c}_{i}$ is replaced by Gaussian's depth $z_i$ and normal $\mathbf{n}_{i}$.

\subsection{Progressive Gaussian Propagation}
\label{subsec:propagation}

\begin{figure}
\centering    
\includegraphics[width=\linewidth]{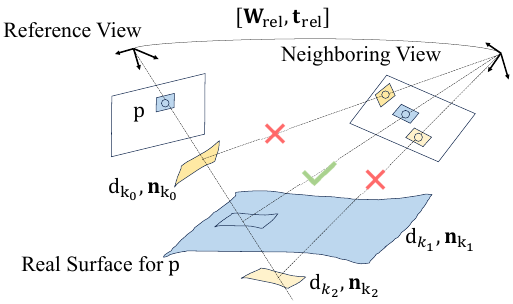}
\caption{Patch matching. To select the best plane candidate for pixel $p$ during propagation, we perform homography transformation between $p$ and each plane candidate, thus yielding the possible corresponding pixels of the neighboring view.
The plane candidate that exhibits the highest color consistency between $p$ and its possible paired pixel is chosen to be the solution. The chosen plane candidate is used to update the depth and normal of pixel $p$.} 
\label{fig: homography}
\vspace{-0.3cm}
\end{figure}
\label{sec:propagation}

In this section, we introduce the progressive Gaussian propagation strategy, which 
allows for propagating accurate geometry from well-modeled regions to under-modeled regions, enabling producing new Gaussians. As shown in Figure~\ref{fig: propagation}, with the rendered depth maps and normal maps, we employ patch matching~\cite{barnes2009patchmatch} to propagate the depth and normal information from neighboring pixels to the current pixel, which produces new depths and normals (named as propagated depth/normal). We further perform geometric filtering and selection operations to pick up the pixels that need more Gaussians and leverage their propagated depths and normals to initialize new Gaussians.


\textbf{Plane Definition.}
To achieve the propagation, the depth and normal of each pixel need to be converted to a 3D local plane first.
For each pixel with its coordinate $\mathbf{p}$, the corresponding 3D local plane is parameterized as $(d, \mathbf{n})$, where $\mathbf{n}$ is the pixel's rendered normal, and $d$ is the distance from the origin of the camera coordinate to the local plane calculated as:
\begin{equation}
\label{eq:distance}
d = z \mathbf{n}^{\top} \mathbf{K}^{-1} \widetilde{\mathbf{p}},
\end{equation}
where $\widetilde{\mathbf{p}}$ is the homogeneous coordinate of $\mathbf{p}$, $z$ is pixel's rendered depth, and $\mathbf{K}$ refers to the camera intrinsic. 

\textbf{Candidate Selection.}
After defining the 3D local plane, the neighbors of each pixel need to be selected for propagation. 
We follow the checkerboard pattern defined in ACMH~\cite{xu2019multi} to select neighboring pixels. For clarity, we illustrate the propagation of a pixel with its four nearest pixels. For each pixel, a set of plane candidates $\left\{\left(d_{k_l}, \mathbf{n}_{k_l}\right) \mid l \in\{0,1,2,3,4\}\right.$ is obtained through propagation ($k_l$ refers to the index of pixel $p$ and its four neighboring pixels). 

\textbf{Patch Matching.}
After obtaining the plane candidates, the optimal plane for each pixel is determined through patch matching.
For a pixel $p$ with its coordinate $\mathbf{p}$, a homography transformation $\mathbf{H}$ is performed based on each plane candidate $\left(d_{k_l}, \mathbf{n}_{k_l}\right)$, which warps $\mathbf{p}$ to $\mathbf{p}^{\prime}$ in the neighboring frame as:
\begin{equation}
\widetilde{\mathbf{p}^{\prime}} \simeq \mathbf{H} \tilde{\mathbf{p}},
\end{equation}
where $\widetilde{\mathbf{p}^{\prime}}$ is the homogeneous coordinate of $\mathbf{p}^{\prime}$, and $\mathbf{H}$ can be induced as:
\begin{equation}
\mathbf{H}=\mathbf{K}\left(\mathbf{W}_{\text {rel}}-\frac{\mathbf{t}_{\text {rel}} \mathbf{n}_{k_l}^{\top}}{d_{k_{l}}}\right) \mathbf{K}^{-1},
\end{equation}
where $[\mathbf{W}_{\text{rel}}, \mathbf{t}_{\text{rel}}]$ is the relative transformation from the reference view to the neighboring view. Finally, the color consistency of $p$ and $p^{\prime}$ is evaluated based on NCC (Normalized Cross Correlation)~\cite{yoo2009fast}. 
The local plane of $p$ will be updated to the plane candidate with the best color consistency. 
Fig.~\ref{fig: homography} also provides an intuitive visualization of this process.
The propagation for plane candidates is iterated $u$ times to transmit effective geometric information over a large region. Then the pixel's depth and normal are updated from the propagated plane, ultimately resulting in the propagated depth and normal maps in Fig.~\ref{fig: propagation}.

\textbf{Geometric Filtering and Selection.}
Due to the inevitable errors in the propagated results, we filter out inaccurate depth and normal through multi-view geometric consistency check~\cite{schonberger2016pixelwise} and obtain filtered depth and normal maps. Finally, we calculate the absolute relative difference between the filtered depth and rendered depth. For regions with an absolute relative difference greater than the threshold $\sigma$, we consider that existing Gaussians fail to model these regions accurately. Therefore, we project pixels in these regions back to the 3D space and initialize them as 3D Gaussians using the same initialization in 3DGS. These Gaussians are then added to the existing Gaussians for further optimization.

\subsection{Plane Constraint Optimization}
\label{sec:normal}
\begin{figure}
\centering    
\includegraphics[width=\linewidth]{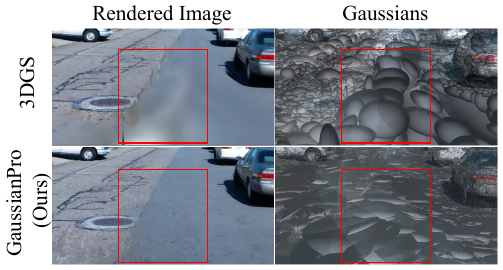}
\caption{Visual comparisons with 3DGS on novel view synthesis. The rendered image of 3DGS contains severe artifacts since the Gaussian spheres are out of order and do not accurately model the true geometry.
On the contrary, our method faithfully captures the details of the road, and its Gaussian spheres are more compact and orderly.} 
\label{fig: normal}
\vspace{-0.2cm}
\end{figure}

In the original 3DGS, the optimization only relies on image reconstruction loss without incorporating any geometric constraints. As a result, the optimized Gaussian shapes may deviate significantly from the actual surface geometry. This deviation leads to a decline in the rendering quality when viewed from a new viewpoint, particularly for large-scale scenes with limited views. As shown in Fig.~\ref{fig: normal}, the shape of Gaussians in 3DGS differs significantly from the road's geometry, resulting in severe rendering artifacts when viewed from a novel viewpoint.
In this section, we propose a planar constraint that encourages the shape of Gaussians to closely resemble the real surface.

Specifically, the propagated 2D normal map in Section~\ref{sec:propagation} represents the orientation of the planes in the scene. We explicitly enforce the consistency between Gaussian's rendered normal and the propagated normal with $\mathcal{L}_1$ and angular loss as $\mathcal{L}_{\text{normal}}$:
\begin{equation}
\mathcal{L}_{\text {normal}}=\sum_{\mathbf{p} \in \mathcal{Q}}\|\hat{N}(\mathbf{p})-\bar{N}(\mathbf{p})\|_1+\left\|1-\hat{N}(\mathbf{p})^{\top} \bar{N}(\mathbf{p})\right\|_1,
\end{equation}

where $\hat{N}$ is the rendered normal map, $\bar{N}$ is the propagated normal map, and $\mathcal{Q}$ refers to the set of valid pixels after the geometric filtering in Section~\ref{sec:propagation}.

Additionally, to ensure that the shortest axis of the Gaussian could represent the normal direction, we incorporate a scale regularization loss $\mathcal{L}_{\text {scale}}$ in NeuSG~\cite{chen2023neusg}. This loss constrains the minimum scale in Gaussian to be close to zero, effectively flattening the Gaussians towards a planar shape.
Finally, the plane constraint can be expressed as the weighted sum of two losses:
\begin{equation}
\mathcal{L}_{\text {planar}}= \beta \mathcal{L}_{\text {normal}} + \gamma  \mathcal{L}_{\text {scale}}.
\end{equation}

\subsection{Training Strategy}
\label{sec:training}
In summary, we incorporate the progressive Gaussian propagation strategy into 3DGS, activating it every $m$ iterations in the optimization, where we set $m=50$. 
The propagated normal maps are saved for computing the planar constraint loss. 
Our final training loss $\mathcal{L}$ consists of the image reconstruction loss $\mathcal{L}_{1}$ and $\mathcal{L}_{\text{D-SSIM}}$ in 3DGS with the proposed planar constraint loss, as illustrated in Eq.~\ref{finalloss}.

\begin{equation}
\label{finalloss}
\mathcal{L}=(1-\lambda) \mathcal{L}_1+\lambda \mathcal{L}_{\text {D-SSIM }} + \mathcal{L}_{\text {planar}},
\end{equation}
where the weight $\lambda$ is set to $0.2$ same with 3DGS.

\vspace{-0.2cm}
\section{Experiment}

\begin{table*}
  \centering
  \footnotesize
  \caption{Quantitative comparisons on Waymo and MipNeRF-360 datasets. We indicate the best and second best with bold and underlined respectively. 3DGS* refers to the results obtained by 3DGS retrained with better SfM point clouds.
  }
  \vspace{-0.1cm}
    \newcolumntype{"}{@{\hskip\tabcolsep\vrule width 1.2pt\hskip\tabcolsep}}
  \label{tab:maincomparison}
  \begin{tabular}{@{}l|c|ccc"ccc@{}}
    \toprule
    & & & Waymo & & & MipNeRF 360 & \\
    Method & FPS $\uparrow$ & PSNR $\uparrow$ & SSIM $\uparrow$ & LPIPS $\downarrow$ & PSNR $\uparrow$ & SSIM $\uparrow$ & LPIPS $\downarrow$ \\
  \hline
    Instant-NGP~\cite{muller2022instant} & 3 & 30.98 & 0.886 & 0.281 & 25.59 & 0.699 & 0.331 \\
    Mip-NeRF 360~\cite{barron2022mip} & 0.02 & 30.09 & 0.909 & 0.262 & 27.69 & 0.792 & 0.237 \\
    Zip-NeRF~\cite{barron2023zip} & 0.09 & \underline{34.22} & \underline{0.939} & \underline{0.205} & \textbf{28.54} & \textbf{0.828} & \textbf{0.189} \\ 
    \hline
    3DGS*~\cite{kerbl20233d} & \underline{103} & 33.53 & 0.938 & 0.226 & 27.21 & 0.815 & 0.214 \\
    3DGS (Retrained) & 102 & - & - & - & 27.88 & 0.824 & 0.209  \\
    GaussianPro (Ours) & \textbf{108} & \textbf{34.68} & \textbf{0.949} & \textbf{0.191} & \underline{27.92} & \underline{0.825} & \underline{0.208} \\
    \bottomrule
  \end{tabular}
\end{table*}

\subsection{Datasets and Implementation Details}
\textbf{Datasets.} We conduct our experiments in a large-scale urban dataset Waymo~\cite{sun2020scalability}, and the common NeRF benchmark Mip-NeRF360 dataset.~\cite{caesar2020nuscenes}. 
On the Waymo dataset, we randomly select nine scenes for evaluation.
To evaluate the performance of novel view synthesis, following the common settings, we select one of every eight images as testing images and the remaining ones as training data. 
We apply the three widely-used metrics for evaluation, $i.e.$, peak signal-to-noise ratio (PSNR), structural similarity index measure (SSIM), and the learned perceptual image patch similarity (LPIPS)~\cite{zhang2018unreasonable}.

\textbf{Implementation Details.}
Our method is built upon the popular open-source 3DGS code base~\cite{kerbl20233d}. In alignment with the approach described in 3DGS, our models are trained for 30,000 iterations across all scenes following 3DGS's training schedule and hyperparameters.
Besides the original clone and split Gaussian densification strategies used in 3DGS, We additionally perform our proposed progressive propagation strategy every $50$ training iterations where propagation is performed 3 times. The threshold $\sigma$ of the absolute relative difference is set to $0.8$. For the planar loss, we set $\beta=0.001$ and $\gamma=100$. All experiments are conducted on an RTX 3090 GPU.


\subsection{Quantative and Qualitative Results}

\begin{figure*}
\centering    
\includegraphics[width=\linewidth]{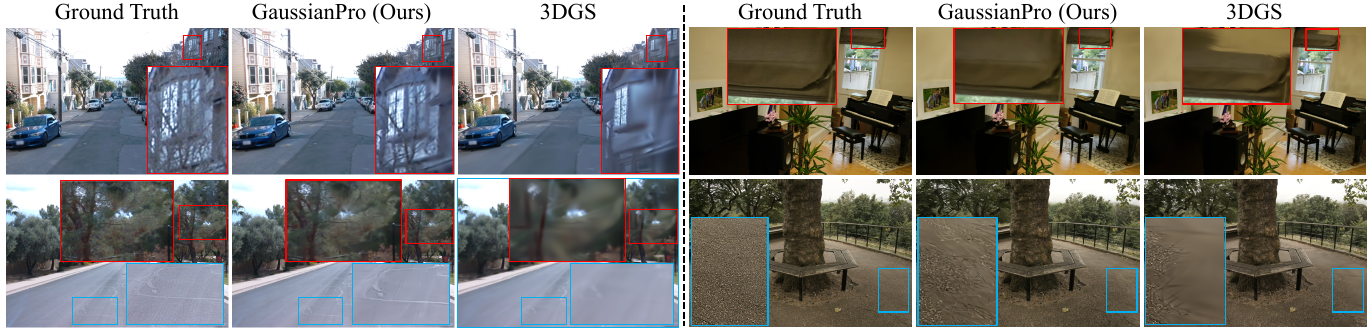}
\caption{Rendering results on the Waymo (left) and MipNeRF360 (right) datasets.  Compared to 3DGS, we have achieved a noticeable improvement in both texture-less surfaces and sharp details.} 
\label{fig:maincompare}
\end{figure*}

As shown in Tab.~\ref{tab:maincomparison}, we compare our method with the state-of-the-art (SOTA) methods, including Instant-NGP~\cite{muller2022instant}, Mip-NeRF360~\cite{barron2022mip}, ZipNeRF~\cite{barron2023zip}, and 3DGS~\cite{kerbl20233d}.

\textbf{Results on Waymo.}  On the large-scale urban dataset Waymo, our method significantly outperforms others in all evaluation metrics. Due to the presence of textureless regions in street views, initializing point clouds in these regions becomes a challenge for SfM. Consequently, it is difficult for 3DGS to densify Gaussians that accurately represent the geometry of the scene in these regions. Otherwise, our propagation strategy accurately complements the missing geometry in the scene. Additionally, our planar constraint allows for better modeling of the scene's planes. Therefore, compared to the baseline 3DGS, our method significantly improves PSNR by 1.15 dB. The visual results presented in Fig.~\ref{fig:maincompare} show that our method achieves sharp details and better renderings in both rich-texture and texture-less regions.

\textbf{Results on MipNeRF360.} On the MipNeRF360, we retrain 3DGS using our generated SfM point clouds since we observed the SfM points used in their official code base can be improved. We report the quantitative results of the original 3DGS (denoted as 3DGS*) and our retrained 3DGS in Tab.~\ref{tab:maincomparison}.
Our method achieves comparable results with 3DGS with a slight improvement.
The MipNeRF360 dataset contains quite small-scale natural and indoor scenes with rich textures, so the SfM techniques usually provide a high-quality point cloud for initialization and the simple clone and split densification strategies don't show a bottleneck in the small-scale scenes.
For indoor scenes with some weak-texture surfaces, our method still shows improvement.
We report results for each scene under MipNeRF360 in the appendix to further support our conclusions.
As shown in Fig~\ref{fig:maincompare}, compared to 3DGS, our method achieves more accurate renderings and clear details.


\subsection{Ablation Study}

\begin{table}
\centering
\footnotesize
\caption{Ablation study on the proposed propagation strategy and planar constraint.}
\label{tab:mainablation}
\begin{tabular}{@{}cc|lll@{}}
\toprule
Propagation & \multicolumn{1}{c}{Planar} & PSNR $\uparrow$ & SSIM $\uparrow$ & LPIPS $\downarrow$ \\ \midrule
 \XSolidBrush  & \XSolidBrush  
 & 33.53 & 0.938 & 0.226 \\
 \XSolidBrush  &  \Checkmark 
 & 34.02 & 0.942 & 0.218 \\
  \Checkmark & \XSolidBrush
 & 34.48 & 0.946 & 0.203   \\
 \Checkmark  & \Checkmark & 
 \textbf{34.68} & \textbf{0.949} & \textbf{0.191} \\ \bottomrule
\end{tabular}
\vspace{-0.5cm}
\end{table}

\begin{figure}
\centering    
\includegraphics[width=\linewidth]{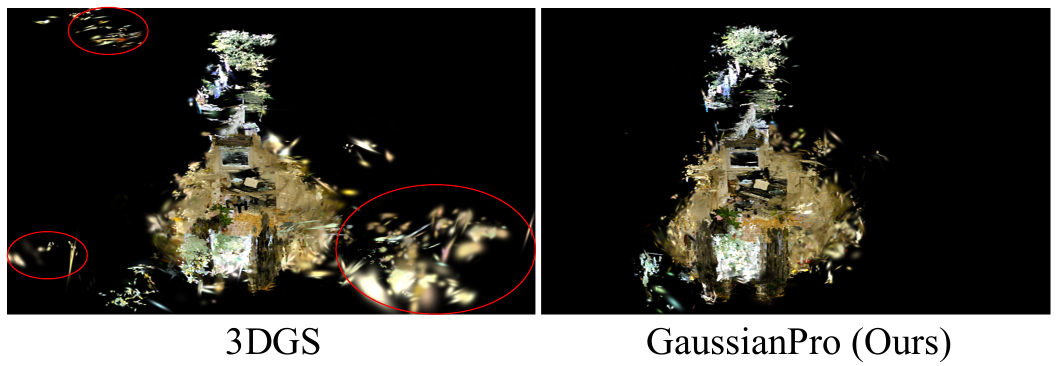}
\caption{Visualization of Gaussians in the \textit{Room} scene of MipNeRF360 dataset. Our method contains fewer noisy Gaussians and achieves a more compact representation.} 
\label{fig: roomgaussians}
\vspace{-0.6cm}
\end{figure}

\begin{figure*}
\centering    
\includegraphics[width=\linewidth]{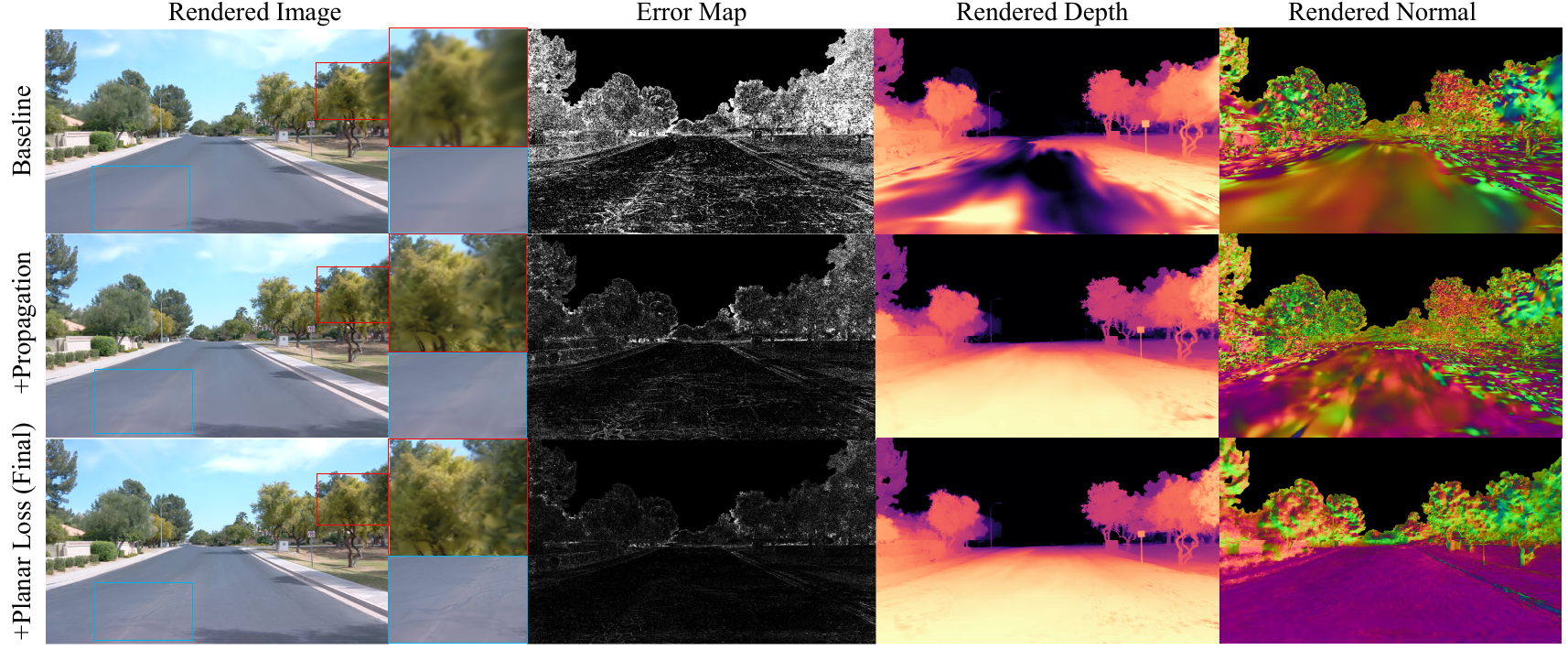}
\caption{The progressive propagation strategy effectively enhances the geometry of the scene, resulting in improved rendering quality. The planar constraint further improves the geometry and rendering of planes.} 
\label{fig: ablation}
\end{figure*}

\textbf{Effectiness of the Propagation Strategy and Planar Constraint.}
We validate the effectiveness of the proposed propagation strategy and planar constraint in the Waymo dataset. As shown in Tab.~\ref{tab:mainablation}, the progressive propagation strategy (the third row) brings significant improvement compared with the baseline. This improvement can be attributed to its ability to refine the geometric representation of the scene, particularly in regions where the initial 3DGS exhibits significant errors (shown in the first and second rows of Fig.~\ref{fig: ablation}).
The planar constraint can further enhance the rendering quality by accurately modeling the normals of the planes, as shown in the third row of Fig.~\ref{fig: ablation}.

\textbf{The Robustness against Sparse Training Images.}
As the number of training images decreases, the rendering quality of neural rendering methods, including 3DGS, tends to decline. 
In Table~\ref{tab:sparsity}, we present the results of training 3DGS and our method using randomly selected subsets comprising $30\%$, $50\%$, $70\%$, $100\%$ of the training images from a scene in MipNeRF360. Remarkably, our method consistently achieves superior rendering results compared to 3DGS across different percentages of training images.

\textbf{Efficiency Analysis.}
We select two typical outdoor and indoor scenes to compare the efficiency of our method with 3DGS, as shown in Tab.~\ref{tab:efficiency}. We achieve a noticeable improvement in rendering quality with a slight increase in training time. In the case of the street scene, 3DGS uses large incorrect Gaussians to represent the ground, as shown in the blue circle of Fig.~\ref{fig:teaser}, resulting in fewer Gaussians compared to our method. However, for the room scene, our method results in more compact Gaussians with less noise (also shown in Fig.~\ref{fig: roomgaussians}). Additionally, our method achieves a comparable real-time rendering frame rate as 3DGS.

\begin{table*}[]
\centering
\footnotesize
\caption{Comparison of 3DGS and ours with different training view ratios in the \textit{room} scene of the MipNeRF360 dataset.}
\label{tab:sparsity}
\begin{tabular}{l|ccc|ccc|ccc|ccc}
\hline
\multirow{2}{*}{Method} & \multicolumn{3}{c|}{30\%}                         & \multicolumn{3}{c|}{50\%}                         & \multicolumn{3}{c|}{70\%}                         & \multicolumn{3}{c}{100\%}                          \\
                        & PSNR           & SSIM           & LPIPS          & PSNR           & SSIM           & LPIPS          & PSNR           & SSIM           & LPIPS          & PSNR           & SSIM           & LPIPS          \\ \hline
3DGS                    & 28.45          & 0.896          & 0.216          & 29.97          & 0.912          & 0.203          & 30.87          & 0.921          & 0.194          & 31.71          & 0.919          & \textbf{0.192}          \\
GaussianPro(Ours)       & \textbf{28.64} & \textbf{0.900} & \textbf{0.210} & \textbf{30.27} & \textbf{0.914} & \textbf{0.199} & \textbf{30.93} & \textbf{0.924} & \textbf{0.189} & \textbf{31.98} & \textbf{0.927} & \textbf{0.192} \\ \hline
\end{tabular}
\vspace{-0.3cm}
\end{table*}

\begin{table}[]
\centering
\scriptsize
\caption{Efficiency analysis. We analyze the effects of initialization using SfM points or MVS points.}
\label{tab:efficiency}
\begin{tabular}{l|c|cccc}
\hline
Scene & Strategy & \multicolumn{1}{c}{PSNR} & \multicolumn{1}{c}{Gaussians} & \multicolumn{1}{c}{Training} & \multicolumn{1}{c}{FPS} \\ \hline
\multirow{3}{*}{Street}  
& SfM points+3DGS & 35.05 & \textbf{665k} & \textbf{40min} & \textbf{119} \\
& MVS points+3DGS & \textbf{36.13} & 1705k & 250min & 75  \\
& SfM points+GaussianPro   & \underline{36.08} & \underline{991k} & \underline{56min} & \underline{108} \\ \hline
\multirow{3}{*}{Room}
& SfM points+3DGS & 31.71 & 1537k & \textbf{59min} & 105 \\
& MVS points+3DGS & \textbf{32.05} & 1832k & 270min & 90 \\
& SfM points+GaussianPro & \underline{31.98} & \textbf{1461k} & \underline{70min} & \textbf{113} \\ \hline
\end{tabular}
\vspace{-0.3cm}
\end{table}

\textbf{Comparison to MVS Inputs.} As our method achieves better rendering quality by improving Gaussians' geometry, it raises the question of whether a similar effect can be achieved by directly inputting denser and more accurate MVS point clouds into 3DGS. To investigate this, we compare the results of optimizing 3DGS with the dense point cloud generated by the MVS method~\cite{schonberger2016pixelwise}. Tab.~\ref{tab:efficiency} shows that directly inputting the MVS point cloud significantly increases the training time (approximately 4 times) due to the additional MVS process and the large number of initial Gaussians. Moreover, the number of Gaussians increases significantly, and the rendering speed noticeably decreases, despite a slight improvement in rendering quality.
Contrarily, our method achieves a favorable balance between rendering quality and efficiency.

\section{Conclusion}
In this paper, we propose GaussianPro, a novel progressive propagation strategy to guide Gaussian densification according to the surface structure of the scene. 
Based on the propagation process, we additionally introduce the plane constraints during optimization to encourage the Gaussains to better model planar surfaces. 
Our method demonstrates superior rendering results compared to 3DGS on both Waymo and MipNeRF360 datasets, while maintaining compact Gaussian representations. Our method shows significant improvements in structured scenes and remains robust to variations in the number of training images.
However, our method does not specially model dynamic objects, and will present artifacts on these regions like all the static Gaussian methods.
In the future, the recent dynamic Gaussian techniques can be incorporated into our method as complementary components to handle dynamic objects.


\bibliography{example_paper}
\bibliographystyle{draft}

\newpage
\appendix
\onecolumn



\end{document}